\documentclass{article}

\usepackage{amsmath,amsfonts,bm}

\def\eqref#1{equation~\ref{#1}}

\def\1{\bm{1}}

\def\vh{{\bm{h}}}

\def\vk{{\bm{k}}}

\def\vm{{\bm{m}}}

\def\vp{{\bm{p}}}

\def\vr{{\bm{r}}}

\def\vw{{\bm{w}}}
\def\vx{{\bm{x}}}

\def\vz{{\bm{z}}}

\def\mJ{{\bm{J}}}

\def\mP{{\bm{P}}}

\def\mR{{\bm{R}}}

\def\mW{{\bm{W}}}
\def\mX{{\bm{X}}}

\def\mZ{{\bm{Z}}}

\def\mSigma{{\bm{\Sigma}}}

\DeclareMathAlphabet{\mathsfit}{\encodingdefault}{\sfdefault}{m}{sl}
\SetMathAlphabet{\mathsfit}{bold}{\encodingdefault}{\sfdefault}{bx}{n}

\usepackage{microtype}
\usepackage{graphicx}
\usepackage{subfigure}
\usepackage{booktabs} 

\usepackage{tikz}
\usetikzlibrary{positioning}
\usepackage{pifont}
\newcommand{\cmark}{\ding{51}}%
\newcommand{\xmark}{\ding{55}}%

\usepackage{hyperref}
\usepackage{bbm}

\usepackage[accepted]{icml2021}

\icmltitlerunning{Self Normalizing Flows}

\begin{document}

\twocolumn[
\icmltitle{Self Normalizing Flows}



\icmlsetsymbol{equal}{*}

\begin{icmlauthorlist}
\icmlauthor{T. Anderson Keller}{bosch,uva}
\icmlauthor{Jorn W.T. Peters}{bosch,uva}
\icmlauthor{Priyank Jaini}{bosch,uva}
\icmlauthor{Emiel Hoogeboom}{bosch,uva}
\icmlauthor{Patrick Forr\'{e}}{uva}
\icmlauthor{Max Welling}{uva}
\end{icmlauthorlist}

\icmlaffiliation{bosch}{UvA-Bosch Delta Lab}
\icmlaffiliation{uva}{University of Amsterdam, Netherlands}
\icmlcorrespondingauthor{T. Anderson Keller}{t.anderson.keller@gmail.com}

\icmlkeywords{Machine Learning, ICML, Normalizing Flows}

\vskip 0.3in
]
\printAffiliationsAndNotice{}



\begin{abstract}
Efficient gradient computation of the Jacobian determinant term is a core problem in many machine learning settings, and especially so in the normalizing flow framework. Most proposed flow models therefore either restrict to a function class with easy evaluation of the Jacobian determinant, or an efficient estimator thereof. However, these restrictions limit the performance of such density models, frequently requiring significant depth to reach desired performance levels. In this work, we propose \emph{Self Normalizing Flows}, a flexible framework for training normalizing flows by replacing expensive terms in the gradient by learned approximate inverses at each layer. This reduces the computational complexity of each layer's exact update from $\mathcal{O}(D^3)$ to $\mathcal{O}(D^2)$, allowing for the training of flow architectures which were otherwise computationally infeasible, while also providing efficient sampling. We show experimentally that such models are remarkably stable and optimize to similar data likelihood values as their exact gradient counterparts, while training more quickly and surpassing the performance of  functionally constrained counterparts. 
\end{abstract}

\section{Introduction}
\label{sec:intro}

The framework of normalizing flows \cite{tabak2013family} allows for powerful exact density estimation through the change of variables formula \cite{rudin1987real}. A significant challenge with this approach is the Jacobian determinant in the objective, which is generally expensive to compute. A large body of work has therefore focused on methods to evaluate the Jacobian determinant efficiently, usually by limiting the expressivity of the transformation. Two classes of functions have been proposed to achieve this: i) those with triangular Jacobians, such that the determinant only depends on the diagonal  \cite{bogachev2005triangular, marzouk2016introduction, jaini2019sum}, and ii) those which are Lipschitz continuous such that Jacobian determinant can be approximated at each iteration through an infinite series \cite{behrmann2019iresnet,grathwohl2019ffjord}. 
The drawback of both of these approaches is that they rely on strong functional constraints. 

\begin{figure*}
    \centering
\includegraphics[width=0.96\textwidth]{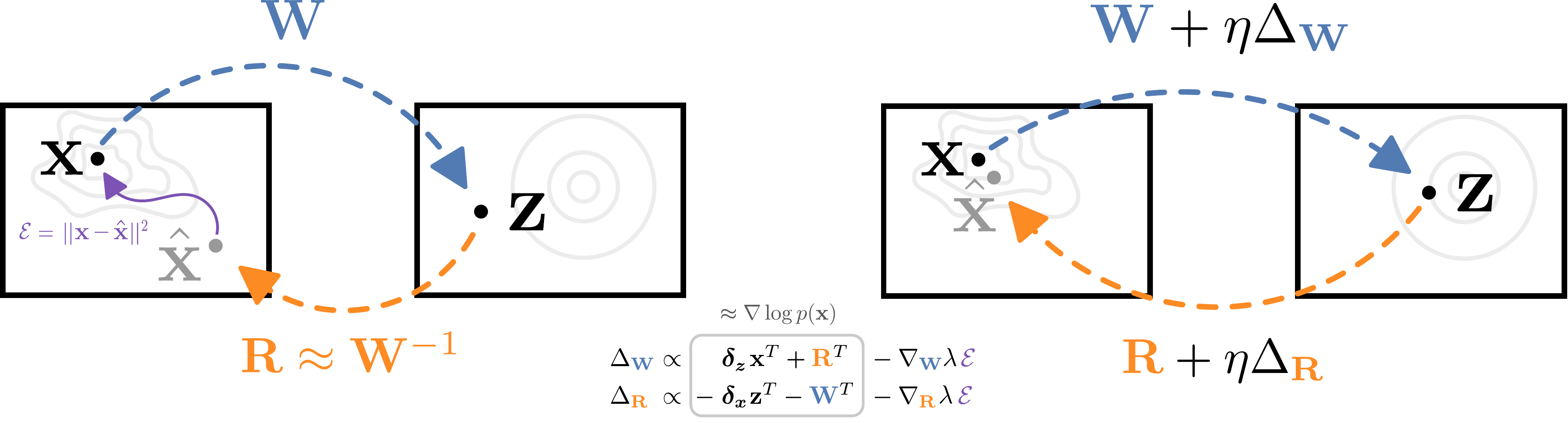}
    \caption{Overview of self normalizing flows. A matrix $\mW$ transforms $\vx$ to $\vz$. The matrix $\mR$ is constrained to approximate the inverse of $\mW$ with a reconstruction loss $\mathcal{E}$. The likelihood is efficiently optimized by approximating the gradient of the log Jacobian determinant with the learned inverse.}
    \label{fig:overview}
\end{figure*}

Recently, a number of works have pursued an alternative approach to training unconstrained normalizing flows by avoiding the expensive computation of the Jacobian determinant altogether during training \cite{gresele2020relative, kramer2020}.  However, these works restrict the parametrization of the transformation, or the parameter updates. As a result, it can be difficult to scale these to higher dimensional data such as images.


In this work we introduce a new framework to avoid the determinant computation during training, which we name the \textit{Self Normalizing Framework}. Instead of computing the log Jacobian determinant, we approximate its gradient directly. This can be achieved through the insight that the derivative of the log Jacobian determinant is given by the inverse of the Jacobian itself. In the framework, flow components learn to approximate their own inverse through a self-supervised layer-wise reconstruction loss. Further, we then define the overall density model as a mixture of the probability induced by both the forward and inverse transformations and show how both transformations can be updated symmetrically using their respective learned inverses directly in the gradient. Ultimately, this avoids the $\mathcal{O}(D^3)$ complexity required for computing the determinant of each layer at each training iteration, instead substituting it with an additional backwards pass of order $\mathcal{O}(D^2)$ required to propagate the reconstruction error gradients.

\section{Related Work}
The field of normalizing flows can be broadly divided into linear and non-linear flows \citep{papamakarios2019normalizing}. Non-linear flows are generally constructed either by constraining the Jacobian to be triangular \citep{kingma2016iaf,maf,berg2018sylvester,huang2018neuralautoregressiveflow,jaini2019sum} or by constraining a residual function to be Lipschitz \citep{grathwohl2019ffjord,behrmann2019iresnet,chen2019residual,perugachi2020idensenets}. Linear flows are constructed using a variety of methods. Examples of linear flows are $1 \times 1$ convolutions \citep{kingma2018glow} which have block-diagonal structure, periodic convolutions \citep{karami2019conv,hoogeboom2019emerging} which leverage the frequency domain, Woodbury flows \cite{lu2020woodbury} that use low-rank transformations and relative gradient-based flows \citep{gresele2020relative} that re-frame optimization of fully connected linear flows. The disadvantage of these methods is that they are either constrained to a subset of transformations, or based on matrix decomposition structures that cannot straightforwardly be extended to convolutional weight sharing.


The work of \citet{gresele2020relative} is most similar to ours in the goal of training unconstrained normalizing flows through the use of efficient gradient computations. In \citet{gresele2020relative} this is achieved by applying a carefully constructed post-conditioner to the gradient of each layer, transforming it into the natural gradient \cite{rel_grad_orig, amari_naturalgrad}, thereby selectively canceling out the inverse normally required during training. However, since parameters are extensively shared for convolutions, natural gradient methods cannot be straightforwardly applied to convolutional layers. In contrast, the framework proposed in this paper makes use of the traditional gradient, allowing for more flexibility in parameterization, requiring only that an inverse function can be learned and maintained throughout training.

Related to our framework, the idea of using learned inverse functions in the setting of density estimation was proposed in early work on invertible neural networks \cite{rippel2013}. In that work, similar to ours, both directions of the density model are parameterized and constrained to be approximate inverses through a reconstruction loss. The learned `encoder' is then used to approximate the marginal distribution of the data in latent space by finding the best fit of a tractable parametric family (such as the Beta family), and the divergence of the approximate latent distribution from the target latent distribution is then minimized. Important differences with our work are that we compute the likelihood exactly through the change of variables formula, and use the Jacobian of our learned `decoder' as an approximation to the gradient of the intractable Jacobian determinant term which arises. An advantage of this approach is that we no longer need to explicitly constrain our decoder to be well conditioned, or invertible, since this constraint is implicitly imposed through the maximization of the Jacobian determinant. Additionally, our framework novelly models the density of observations as a mixture of the density under the forward and inverse transformations, taking advantage of all learned parameters.



More broadly, the idea of greedy layer-wise learning has been explored in many forms for training neural networks \cite{bengio_greedy, hinton_dbn, greedy_infomax}. One influential class of work uses stacked auto-encoders, or deep belief networks, for pre-training or representation learning \cite{bengio_greedy, hinton_dbn, vincent08, scae}. Our work leverages similar models and training paradigms, but introduces them into a modern flow framework, demonstrating additional potential uses for the learned feedback connections.

\begin{figure*}[h!]
\centering
  \includegraphics[width=0.96\textwidth]{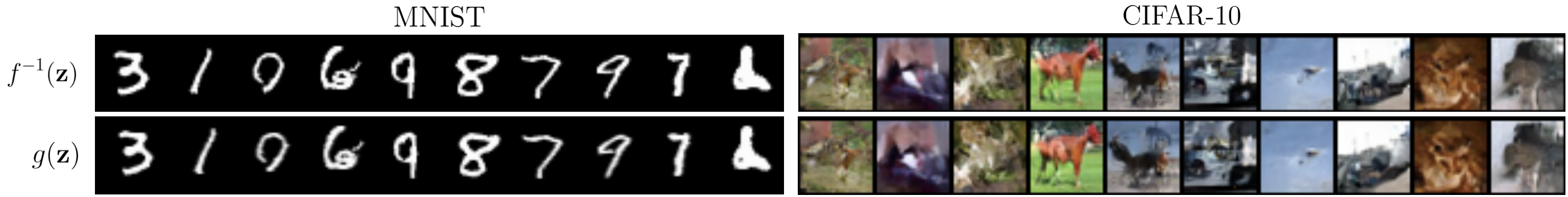}%
  \caption{MNIST (left) and CIFAR-10 (right) samples generated from a self normalizing Glow model using the exact inverse (top) vs. the approximate learned inverse (bottom). From these samples we observe that the self normalizing flow models have learned to become good generative models of the data. Additionally, comparing the top and bottom rows, we see the inverse approximation is nearly exact.}%
  \label{fig:samples}
\end{figure*}

Another related class of work addresses the biological implausibility of backpropagation also through learned layer-wise autoencoders. Target propagation \cite{lecun86, dtp14, dtp15, dtp20, meulemans2020theoretical} addresses the so-called weight transport problem by training auto-encoders at each layer of a network and using these learned feedback weights to propagate `targets' to previous layers. Our method takes inspiration from this approach. Specifically, our method can be viewed as a hybrid of target propagation and backpropagation \cite{Linnainmaa1976, werbos, Rumelhart1986} particularly suited to unsupervised density estimation in the normalizing flow framework. The novelty of our approach in this regard lies in the use of the inverse weights directly in the update, rather than in the backward propagation of updates. 

\section{A General Framework for Self Normalizing Flows}
\label{sec:general_framework}
\subsection{Preliminaries}
Given an observation $\vx \in \mathbb{R}^{D}$, it is assumed that $\vx$ is generated from an underlying real vector $\vz \in \mathbb{R}^D$ through an invertible and differentiable transformation $g$, and that $f = g^{-1}$ is also differentiable (i.e. $g$ is a diffeomorphism). It is further assumed that $\vz$ is a sample from a simple known underlying distribution $p_{\mZ}$ such as a standard Gaussian. Then, the probability density $p_{\mX}$ can be computed exactly using the change of variables formula:
\begin{align}
    \begin{split}
    \label{cov}
    p_{\mX}(\vx) = p_{\mZ}(\vz) \left|\frac{\partial \vz}{\partial\vx}\right| & = p_{\mZ}\left(g^{-1}(\vx)\right) \left|\mJ_{g^{-1}}\right| \\ & = p_{\mZ}\big(f(\vx)\big) \left|\mJ_f\right|
    \end{split}
\end{align}
where the change of volume term $\left| \mJ_f \right| = \left|\frac{\partial f(\vx)}{\partial\vx}\right|$ is the determinant of the Jacobian of the transformation between $\vz$ and $\vx$, evaluated at $\vx$. 

Typically, the functions $f$ and $g$ are defined as compositions of diffeomorphisms themselves, i.e. $g = g_0 \circ g_1 \circ \dots g_K$ and $f = f_K \circ f_{K-1} \circ \dots f_0$ where ${g_k}^{-1} = f_k$. This formulation takes advantage of the fact that a composition of diffeomorphic functions is also a diffeomorphism, meaning that if each $f_k$ is invertible and differentiable, then so is the composition $f$ and the change of variables formula in Equation \ref{cov} still holds. 

Most approaches then propose defining and parameterizing only one direction of the flow, for example the 'forward' functions $f_k$, and compute the inverses $g_k$ exactly when needed. The log-likelihood of the observations is then simultaneously maximized, with respect to a given $f_k$'s vector of parameters $\bm\theta_k$, for all $k$, requiring the gradient. Using the identity $\frac{\partial}{\partial \mJ} \log |\mJ| = {\mJ^{-T}}$ we obtain the following gradient of the loss with respect to a given layer $k$'s parameters: 
\begin{equation}
    \label{grad_cov}
   \text{\scalebox{0.85}{$\frac{\partial}{\partial  \bm\theta_k} \log p_{\mX}(\vx) = \frac{\partial}{\partial \bm\theta_k} \log p_{\mZ}(f(\vx)) + \frac{\partial\left(\mathrm{vec}\ \mJ_{f}\right)^T}{\partial \bm\theta_k} \left( \mathrm{vec}\ \mJ_{f}^{-T} \right)$}}
\end{equation}
Following the conventions of \citet{matrix_derivative} for matrix derivatives, we make use of the vectorization operator $\mathrm{vec}$ which maps from $m \times n$ matricies to $mn \times 1$ column vectors by stacking the columns of the matrix sequentially, and formulate the parameters $\bm\theta_k$ as column vectors.


\subsection{Self Normalizing Framework}
In order to avoid the inverse Jacobian in the gradient, we instead propose to define and parameterize \emph{both} the forward and inverse functions $f_k$ and $g_k$ with parameters $\bm\theta_k$ and $\bm\gamma_k$ respectively. We then constrain the parameterized inverse $g_k$ to be approximately equal to the true inverse $f^{-1}_k$ through a layer-wise reconstruction loss. We can thus define our maximization objective as the mixture of the log-likelihoods induced by both models minus the reconstruction penalty:
\begin{equation}
\label{mixture}
\text{\scalebox{0.85}{$\mathcal{L}(\vx) = \frac{1}{2} \log p^{f}_{\mX}(\vx) + \frac{1}{2} \log p^{g}_{\mX}(\vx) - \lambda \sum_{k=0}^K ||g_k \left( f_k (\vh_k) \right) - \vh_k ||^2_2$}}
\end{equation}
where $p^{f}_{\mX}$ and $p^{g}_{\mX}$ now denote the densities induced by both the forwards and inverse transformations separately, and $\vh_k = \mathrm{gradient\_stop}(f_{k-1} \circ ... f_0 (\vx))$ is the output of function $f_{k-1}$ with the gradients blocked such that only $g_k$ and $f_k$ receive gradients from the reconstruction loss at layer $k$. We see that when $f = g^{-1}$ exactly, this is equivalent to the traditional normalizing flow framework. 

By the inverse function theorem, we know that the inverse of the Jacobian of an invertible function is given by the Jacobian of the inverse function, i.e. $\mJ_f^{-1}(\vx) = \mJ_{f^{-1}}(\vz)$. Therefore, we see that with the above parameterization and constraint, we can approximate both the change of variables formula, and the gradients for both functions, in terms of the Jacobians of the respective inverse functions. Explicitly:
\begin{equation}
    \label{grad_f}
    \text{\scalebox{0.9}{$
    \frac{\partial}{\partial  \bm\theta_k} \log p^{f}_{\mX}(\vx) \approx \frac{\partial}{\partial \bm\theta_k} \log p_{\mZ}(f(\vx)) + \frac{\partial \left(\mathrm{vec}\ \mJ_{f}\right)^T}{\partial \bm\theta_k}\left(\mathrm{vec}\ \mJ_{g}^T\right)$}}
\end{equation}
\begin{equation}
    \label{grad_g}
    \text{\scalebox{0.9}{$\frac{\partial}{\partial  \bm\gamma_k} \log p^{g}_{\mX}(\vx) \approx \frac{\partial}{\partial \bm\gamma_k} \log p_{\mZ}(g^{-1}(\vx)) - \frac{\partial \left(\mathrm{vec}\ \mJ_{g}\right)^T}{\partial \bm\gamma_k}\left(\mathrm{vec}\ \mJ_{f}^T\right)$}}
\end{equation}
where Equation $\ref{grad_g}$ follows from the derivation of Equation $\ref{grad_f}$ and the application of the derivative of the inverse. We note that the above approximation requires that the Jacobians of the functions are approximately inverses \emph{in addition} to the functions themselves being approximate inverses. For the models presented in this work, this property is obtained for free since the Jacobian of a linear mapping is the matrix representation of the map itself. However, for more complex mappings, this may not be exactly the case and should be constrained explicitly. We suggest this could be efficiently implemented with an additional loss analogous to a reconstruction loss but employing Jacobian vector products instead of matrix vector products (see Section \ref{sec:extensions}).
\begin{figure}
    \centering
    \includegraphics[width=0.48\textwidth]{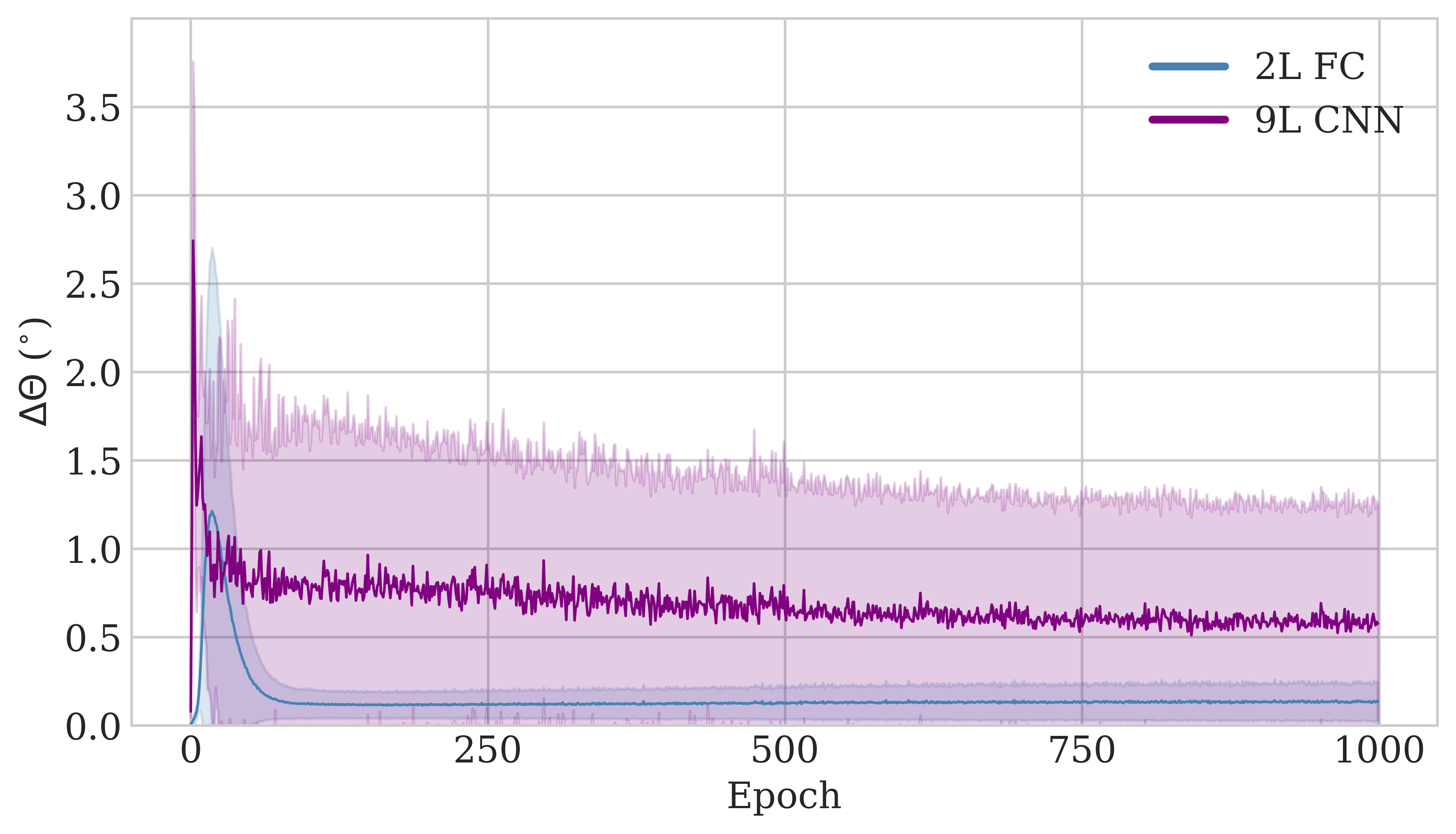}
    \vspace{-8mm}
    \caption{Angle (in degrees) between the exact gradient and the approximate gradient at each step of training. We see that the angle is close to 0.1 degrees for the fully connected network (FC), and less than 1 degree for the convolutional network (CNN), suggesting the approximation is close throughout training.}
    \label{fig:angle}
\end{figure} 

\begin{figure}
\centering
    \includegraphics[width=0.48\textwidth]{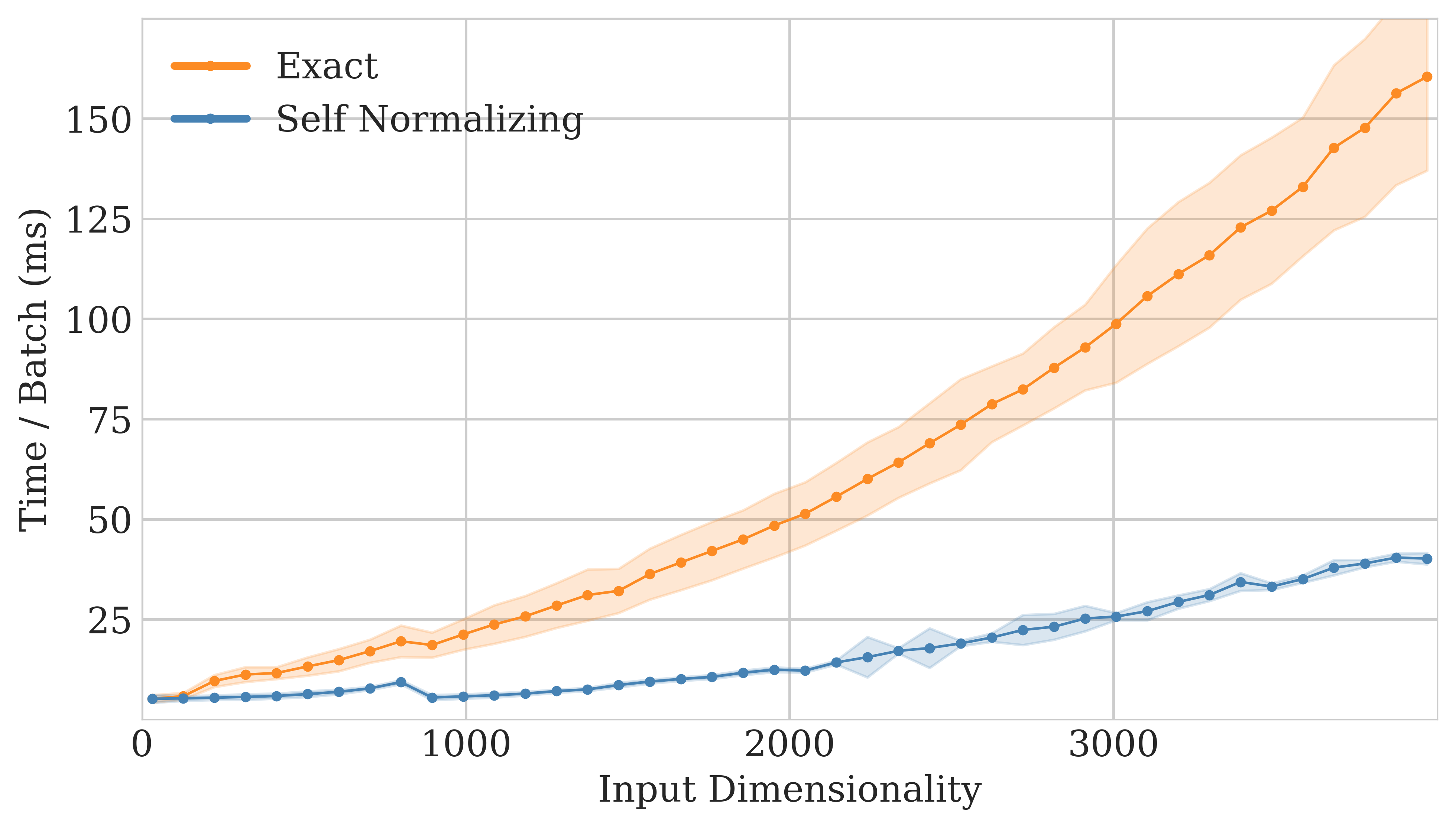}
  \vspace{-7mm}
  \caption{Time per batch vs. input dimensionality for a single fully connected layer, comparing the self normalizing and exact gradients. We see the self normalizing flow model scales much more favorably with input dimension.}
  \label{fig:timing}%
\end{figure}

\looseness=-1
Although there are no known convergence guarantees for such a method, we observe in practice that, with sufficiently large values of $\lambda$, most models quickly converge to solutions which maintain the desired constraint. This phenomenon is demonstrated in Figure \ref{fig:angle} where it is shown that the angle between the true gradient and the approximate gradient at each training iteration decreases to less than 1 degree over the course of training for both models tested. As a visual example of the quality of the inverse approximation, Figure $\ref{fig:samples}$ shows samples from the base distribution $p_{\mZ}$ passed through both the true inverse $f^{-1}$ (top) and the learned approximate inverse $g$ (bottom) to generate samples from $p_{\mX}$. As demonstrated by the nearly identical samples, the approximate inverse appears to be a very close match to the true inverse. The details of the models which generated these samples are in Section \ref{sec:exp}. Further mitigation strategies for potential optimization difficulties are discussed in Sections \ref{sec:discussion} and \ref{sec:extensions}. Finally, in Figure \ref{fig:timing} we see the efficiency improvements of our method compared with the exact gradient. As expected, 
we see the self normalizing method scales much more favorably with input dimensionality than the exact gradient method.

\subsection{Inference with Self Normalizing Flows}
\label{sec:inference}
The above framework proposes approximate gradients which allow for the training of normalizing flow architectures without having to compute expensive determinants or matrix inverses. However, to compute the exact log-likelihood of an observation (i.e. perform inference), the exact log Jacobian determinant still has to be computed for each input, which may be expensive.

\begin{figure*}
    \centering
    \includegraphics[width=1.0\textwidth]{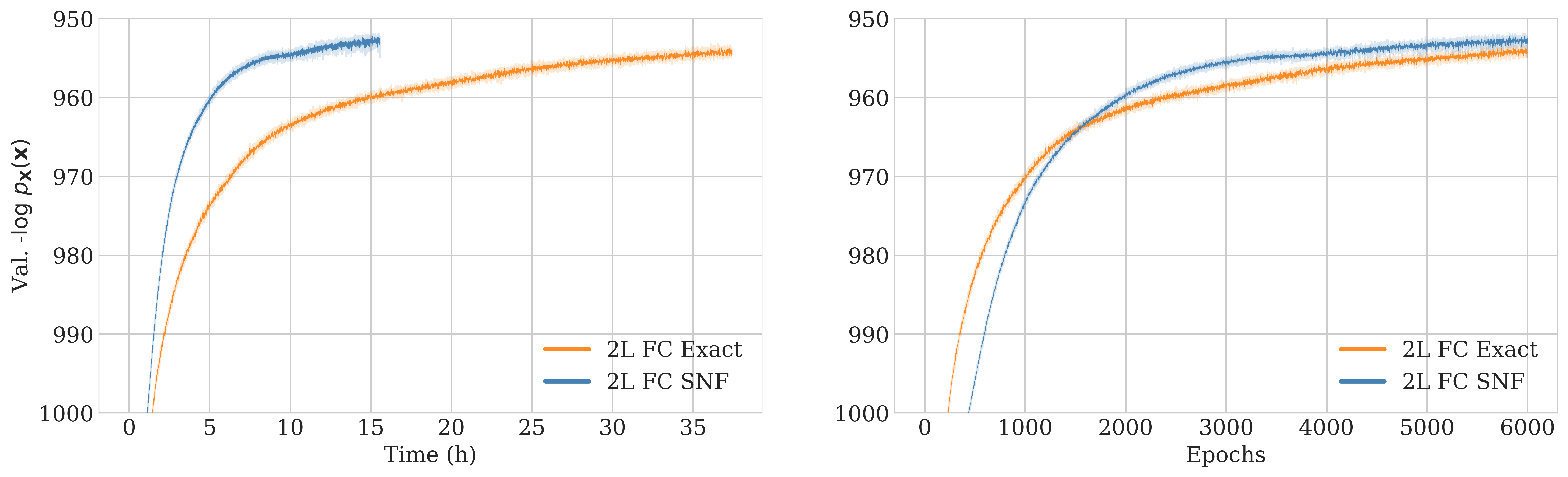}
    \caption{Negative log-likelihood on the MNIST validation set for a  2-layer fully connected flow trained with exact vs. self normalizing (SNF) gradients. Shown vs. training time (left) and vs. epochs (right). We see that both models converge to similar optima while the SNF model trains much more quickly.}
    \label{fig:training_curve_2L}
\end{figure*}
To alleviate this limitation we take advantage of the compositional formulation of $f$, and the multiplicative property of the determinant. In detail, the determinant of the product of square matrices is equal to the product of their determinants. Therefore, assuming our flow is composed of square transformations $f_k$, the Jacobian is given by $\mJ_f = \prod_k \mJ_{f_k}$, and the log determinant of the Jacobian is given by:
\begin{equation}
    \label{det_fac}
    \log |\mJ_f| = \sum_k^K \log |\mJ_{f_k}|
\end{equation}
For neural network architectures composed of sequential linear and nonlinear transformations, this allows us to separate the components of the Jacobian which are data-independent (e.g. those of linear layers) from those that are data-dependant (e.g. those of the activations). For example, given a network composed of $L$ layers of the form $\vh_l = \sigma_l(\mW_l \vh_{l-1})$ where $\sigma_l$ is a point-wise non-linearity, combining Equations \ref{cov} and \ref{det_fac} yields:
\begin{equation}
    \label{cov_fac}
    \log p_{\mX}(\vx) = \log p_{\mZ}(\vz) + \sum_k^K \log |\mW_k| + \sum_k^K \log |\mSigma_k(\vx)|
\end{equation}
where $\mSigma_k(\vx)$ is the Jacobian of the activation function at layer $k$, evaluated at the sample $\vx$.  Importantly, we observe that the data-independent terms, $\sum_k \log |\mW_k|$, are the expensive part of inference, and the data-dependent terms, $\sum_k \log |\mSigma_k(\vx)|$, are frequently cheap to compute analytically given the derivative of the activation. Therefore, as can be seen, once trained, the expensive data-independent terms must only be computed once and their values can then be reused for all future examples, effectively amortizing the cost of the determinant computation.

\section{Self Normalizing Flows}
\label{sec:method}
In this section we introduce two simple applications of the self normalizing framework: fully-connected and convolutional self normalizing flows.

\subsection{Self Normalizing Fully Connected Layer}
As a specific case of the self normalizing framework, we introduce a single fully connected self normalizing layer, as exemplified in Figure $\ref{fig:overview}$. Let $\mW,\mR \in \mathbb{R}^{D \times D}$, with $f(\vx) = \mW \vx = \vz$, and $g(\vz) = \mR \vz$, such that $\mW^{-1} \approx \mR$. We additionally denote the layer-wise reconstruction loss as $\mathcal{E}(\vx) = ||\mR \mW \vx - \vx||^2_2$, but leave the derivation of the gradients of this term for the appendix (see Section  \ref{appendix:fc_derivation}) since these are efficiently computed by standard deep learning frameworks and require no approximations.

Taking the gradients of Equation $\ref{mixture}$ with respect to all parameters $\mW$ and $\mR$, 
we get the following exact gradients:
\fontsize{7.8}{10}
\begin{align}
\frac{\partial}{\partial  \mW} \mathcal{L}(\vx)
& = \frac{1}{2}\frac{\partial}{\partial \mW} \log p^{f}_{\mX}(\vx) + \frac{1}{2}\frac{\partial}{\partial \mW} \log p^{g}_{\mX}(\vx) - \lambda  \frac{\partial}{\partial \mW} \mathcal{E} \nonumber \\ 
& = \frac{1}{2}\left(\frac{\partial \log p_{\mZ}(\mW \vx)}{\partial \mW\vx} \frac{\partial \mW \vx}{\partial \mW}  + \frac{\partial \log | \mW |}{\partial \mW}\right)  - \lambda \frac{\partial}{\partial \mW}  \mathcal{E} \nonumber \\ 
& = \frac{1}{2}\left(\bm\delta_{\vz}^{f} \vx^T + \mW^{-T}\right) - \lambda \frac{\partial}{\partial \mW}  \mathcal{E}
\label{fc_grad_w}
\end{align}
\begin{align}
    \frac{\partial}{\partial  \mR} \mathcal{L}(\vx) &  = \frac{1}{2}\frac{\partial}{\partial \mR} \log p^{f}_{\mX}(\vx) + \frac{1}{2}\frac{\partial}{\partial \mR} \log p^{g}_{\mX}(\vx) - \frac{\partial}{\partial \mR} \lambda \mathcal{E} \nonumber \\
    & = \frac{1}{2} \!\! \left(\frac{\partial \log p_{\mZ}(\mR^{-1}\vx)}{\partial \mR^{-1}\vx} \! \frac{\partial \mR^{-1} \vx}{\partial \mR} \!+\! \frac{\partial \log | \mR^{-1} |}{\partial \mR}\right) \!-\! \lambda \frac{\partial}{\partial \mR}  \mathcal{E} \nonumber \\
    & = \frac{1}{2} \!\! \left(-\mR^{-T} \bm\delta_{\vz}^{g} \vx^T \mR^{-T} - \mR^{-T}\right) \!-\! \lambda \frac{\partial}{\partial \mR}  \mathcal{E}
\label{fc_grad_r}
\end{align}
\normalsize
where $\bm\delta^f_{\vz} = \frac{\partial \log p_{\mZ}(\mW \vx)}{\partial \mW\vx}$ and  $\bm\delta^g_{\vz} = \frac{\partial \log p_{\mZ}(\mR^{-1} \vx)}{\partial \mR^{-1} \vx}$ can be computed by standard backpropagation. To avoid computing matrix inverses, using our framework we substitute $\mW^{-1}$ with $\mR$ in Equation \ref{fc_grad_w}, and symmetrically, all instances of $\mR^{-1}$ with $\mW$ in Equation \ref{fc_grad_r}. Additionally, we approximate $\bm\delta^f_{\vz} \approx \bm\delta^g_{\vz}$ in Equation \ref{fc_grad_r} such that we do not need to `forward propagate' through the inverse model, and can re-use the backpropagated error from the forward model. This corresponds to assuming that $\mR^{-1}\vx \approx \mW\vx$ for all $\vx$ which follows directly from $\mR^{-1} \approx \mW$. Ultimately this results in the following approximations to gradient: 
\small
\begin{equation}
    \label{grad_w_approx}
    \frac{\partial}{\partial  \mW} \mathcal{L}(\vx)  \approx \frac{1}{2}\left(\bm\delta_{\vz}^{f} \vx^T + \mR^{T}\right) - \lambda \frac{\partial}{\partial \mW}  \mathcal{E} 
\end{equation}
\begin{equation}
    \label{grad_r_approx}
    \begin{split}
    \frac{\partial}{\partial  \mR} \mathcal{L}(\vx) & \approx 
    \frac{1}{2} \left(-\bm\delta^{f}_{\vx} \vz^T - \mW^{T}\right) - \lambda \frac{\partial}{\partial \mR}  \mathcal{E} 
    \end{split}
\end{equation}
\normalsize

where $\bm\delta^f_{\vx} = \frac{\partial \log p_{\mZ}(\mW \vx)}{\partial \vx} = \mW^T \delta^f_{\vz}$. We see that by using such a self normalizing layer, the gradient of the log-determinant of the Jacobian term, which originally required an expensive matrix inversion at each iteration, is approximately given by the weights of the inverse transformation -- sidestepping computation of both the Jacobian and the inverse. Additionally, we see the $\bm\delta^f_{\vx}$ term required for Equation $\ref{grad_r_approx}$ is already computed by traditional backpropagation, making the update efficient. Finally, we note the above gradients can trivially be extended to compositions of such layers, combined with non-linearities, by substituting the appropriate deltas for each layer, and the corresponding layer inputs and outputs for $\vx$ and $\vz$ respectively.

\subsection{Self Normalizing Convolutional Layer} 
\label{snf_cnn}

To construct a self normalizing convolutional layer, we consider the setting where both the forward transformation $f$, and the inverse transformation $g$ are convolutional with the same kernel size. We note, importantly, that the inverse of a convolution operation is not necessarily another convolution. However, for sufficiently large $\lambda$, we observe that $f$ is regularized such that it is restricted to the class of convolutions which is approximately invertible by a convolution.

We define the parameters of $f$ to be the kernel $\vw$ and similarly, the parameters of $g$ to be the kernel $\vr$. Then, letting $f(\vx) = \vw \star \vx = \vz$ and $g(\vz) = \vr \star \vz$, such that $f^{-1} \approx g$ with $\vx, \vz \in \mathbb{R}^D$, we proceed to derive the exact gradients of the log-likelihood. Again, we ignore the reconstruction term for simplicity as it requires no approximations. 

To make the derivation easier, we note that the convolution operation is a linear operation, and can therefore be represented in matrix form. We define a transformation $\mW = \mathcal{T}(\vw)$, which maps between the convolutional kernel $\vw$ and the corresponding matrix form of the convolution: 
\begin{equation}
    \label{linearconv}
    \vz = \mathcal{T}(\vw) \vx = \vw \star \vx
\end{equation}
Letting $\vw$ be a column vector and again making use of the vectorization operator $\mathrm{vec}$, we compute the exact gradient of the log-likelihood with respect to the  kernel weights $\vw$:
\fontsize{9.0}{10}
\begin{align}
    \frac{\partial}{\partial \vw} & \log p^{f}_{\mX}(\vx) \nonumber \\
    & = \frac{\partial (\mathrm{vec}\ \mathcal{T}(\vw))^T}{\partial \vw} \left(\frac{\partial \log p_{\mZ}\left(\mathcal{T}(\vw) \vx\right)}{\partial\ \mathrm{vec}\ \mathcal{T}(\vw)} + \frac{\partial \log | \mathcal{T}(\vw) |}{\partial\ \mathrm{vec}\ \mathcal{T}(\vw)} \right) \nonumber \\
    & =  \frac{\partial (\mathrm{vec}\ \mathcal{T}(\vw))^T}{\partial \vw} \left(\mathrm{vec}\  \left[\bm\delta^f_{\vz} \vx^T\right] + \mathrm{vec}\ \mathcal{T}(\vw)^{-T}\right) \nonumber \\
    & = \bm\delta^f_{\vz} \star \vx + \frac{\partial (\mathrm{vec}\ \mathcal{T}(\vw))^T}{\partial \vw} \left(\mathrm{vec}\ \mathcal{T}(\vw)^{-T}\right)
    \label{appendix:exact_conv_w}
\end{align}
\normalsize
where we see that the first term $\frac{\partial (\mathrm{vec}\ \mathcal{T}(\vw))^T}{\partial \vw} \left(\mathrm{vec}\  \left[\bm\delta^f_{\vz} \vx^T\right]\right)$ is given by the convolution $\bm\delta^f_{\vz} \star \vx$, as is usually done with backpropagation in convolutional neural networks. Then, given our soft constraint $f^{-1} \approx g$, we can approximate $\mathcal{T}(\vw)^{-T}$ with $\mathcal{T}(\vr)^{T}$ giving us:
\small
\begin{equation}
\label{appendix:approx_conv_w}
\begin{split}
    \frac{\partial}{\partial \vw} \log p^{f}_{\mX}(\vx) & \approx \bm\delta^f_{\vz} \star \vx + \frac{\partial (\mathrm{vec}\ \mathcal{T}(\vw))^T}{\partial \vw} \left(\mathrm{vec}\ \mathcal{T}(\vr)^{T}\right) 
\end{split}
\end{equation}
\normalsize
To simplify the second term we note two points. First, the transpose of a convolution can similarly be achieved by standard convolution with a transformed kernel. We call this transformed kernel $\mathrm{flip}(\vr)$ such that $\mathcal{T}(\mathrm{flip}(\vr)) = \mathcal{T}(\vr)^T$. Succinctly, $\mathrm{flip}(\cdot)$ is implemented by swapping the input and output axes, and mirroring the spatial (height and width) dimensions of the kernel. The second point is that the partial derivative $\frac{\partial (\mathrm{vec}\ \mathcal{T}(\vw))^T}{\partial \vw}$ is given by a rectangular matrix, which, when multiplied by the vectorized form of the convolution matrix $\mathcal{T}(\vw)$ yields a constant multiple $\vm$ element-wise multiplied with the kernel $\vw$. Each multiple element $m_i$ is given by the number of times the associated kernel element $w_i$ is shared across the matrix $\mathcal{T}(\vw)$. We provide the derivation of this more thoroughly in the appendix, as well as an efficient method for calculating $\vm$ for arbitrary convolutions (see Section \ref{appendix:conv_derivation}). In combination, we arrive at the approximate gradient of the likelihood with respect to the kernel $\vw$:
\begin{equation}
    \label{appendix:approx_conv_w_complete}
     \frac{\partial}{\partial \vw} \log p^{f}_{\mX}(\vx)  \approx \bm\delta^f_{\vz} \star \vx + \mathrm{flip}(\vr) \odot \vm\\
\end{equation}


We see that the symmetric derivation can be obtained for the gradient with respect to the kernel $\vr$, as outlined below:
\fontsize{8.0}{10}
\begin{align}
    \frac{\partial}{\partial \vr} & \log p^{g}_{\mX}(\vx) \nonumber \\
    & \!\!\!\!\! = \frac{\partial (\mathrm{vec}\ \mathcal{T}(\vr))^{T}}{\partial \vr} \left(\frac{\partial \log p_{\mZ}\left(\mathcal{T}(\vr)^{-1} \vx\right)}{\partial\ \mathrm{vec}\ \mathcal{T}(\vr)} + \frac{\partial \log | \mathcal{T}(\vr)^{-1} |}{\partial\ \mathrm{vec}\ \mathcal{T}(\vr)} \right) \nonumber \\
    &\!\!\!\!\! =  \frac{\partial (\mathrm{vec}\ \mathcal{T}(\vr))^T}{\partial \vr} \left(\mathrm{vec}\! \left[-\mathcal{T}(\vr)^{-T}\bm\delta^g_{\vz} \vx^T \mathcal{T}(\vr)^{-T}\right] \!-\! \mathrm{vec}\ \mathcal{T}(\vr)^{-T}\right) \nonumber \\
    &\!\!\!\!\! \approx  \frac{\partial (\mathrm{vec}\ \mathcal{T}(\vr))^T}{\partial \vr} \left(\mathrm{vec}\! \left[-\mathcal{T}(\vw)^{T}\bm\delta^f_{\vz} \vx^T \mathcal{T}(\vw)^{T}\right] \!-\! \mathrm{vec}\ \mathcal{T}(\vw)^{T}\right) \nonumber \\
    &\!\!\!\!\! = - \bm\delta^f_{\vx} \star \vz - \mathrm{flip}(\vw) \odot \vm 
\label{appendix:exact_conv_r}
\end{align}
\normalsize

\section{Experiments}
\label{sec:exp}
In our first set of experiments, we train simple flows composed of the above layers and invertible non-linearities on the MNIST dataset. To evaluate our proposed approximate gradients, we compare to baseline models of the same architectures trained with the exact gradient. These architectures are designed to be small, so that it is still possible to compute the exact gradients quickly. We additionally compare with similar recent approaches to training normalizing flows with linear and convolutional layers, namely the relative gradient method of \citet{gresele2020relative} and the convolution parametrizations of  \citet{hoogeboom2019emerging,hoogeboom2020convexp}. 

To evaluate the scalability of our method, we perform a second set of experiments where we integrate self normalizing flows into the Glow framework \cite{kingma2018glow} as a replacement for the 1x1 convolutional mixing layers. In this framework we train models on MNIST, CIFAR-10, and the downsized Imagenet 32x32 dataset. All experimental results are from our re-implementations for consistency. In some cases, due to differing hyper-parameters or errors in prior work, this yielded slightly different results than those published. We provide extended explanations for these discrepancies, as well as a link to our code repository, in the appendix (See Section \ref{appendix:experiment_details}).

\subsection{MNIST}
On the MNIST dataset we train three classes of models: 2-layer fully connected (FC) models with smooth-leaky-ReLU activations \cite{gresele2020relative}, 9-layer convolutional (CNN) models with spline activations \cite{splines}, and 32-layer Glow models composed of affine coupling layers and 1x1 convolutional mixing layers \cite{kingma2018glow}. In all cases, we compare a self normalizing version with its exact gradient baseline.
\begin{table}
    \centering
    \caption{Negative Log-likelihood in nats on the MNIST test set. Mean $\pm$ std. over 3 runs. Self normalizing flows (SNF) achieve comparable performance to their exact counterparts.}
    \vspace{2mm}
    \begin{tabular}{l r}
    \toprule
        Model & $-\log p_{\mX}(\vx)$ \\ \midrule
        Relative Grad. FC 2-Layer  &  1096.5 $\pm$ 0.5 \\
        Exact Gradient FC 2-Layer & 947.6  $\pm$ 0.2\\ 
        SNF FC  2-Layer (ours) & 947.1 $\pm$ 0.2  \\
        \midrule
        Emerging Conv. 9-Layer  &  645.7 $\pm$ 3.6 \\
        SNF Conv. 9-Layer (ours) & 638.6 $\pm$ 0.9 \\
        Conv. Exponential 9-Layer  & 638.1 $\pm$ 1.0 \\ 
        Exact Gradient Conv. 9-Layer &  637.4 $\pm$ 0.2\\
        \midrule
        Glow 2L-16K & 575.7 $\pm$ 0.8 \\
        SNF Glow 2L-16K (ours) & 575.4 $\pm$ 1.4 \\
    \bottomrule
    \end{tabular}
    \label{tab:nll_results}
\end{table}

As can be seen in Table $\ref{tab:nll_results}$ the models composed of self normalizing flow layers are nearly identical in performance to their exact gradient counterparts on the MNIST dataset. We see that the fully connected self normalizing model drastically outperforms the relative gradient method of \citet{gresele2020relative}, reaching the same performance as the exact gradient method. Additionally, we see that the self normalizing convolutional layer outperforms its constrained counterpart from \citet{hoogeboom2019emerging}, likely due to the fact that the emerging convolution is unable to represent the 1x1 convolution explicitly.  We hypothesize that the convolutional self normalizing flow model slightly under-performs the exact gradient method due to the convolutional-inverse constraint. We propose this constraint can be relaxed by using more complex inverse functions, potentially composed of multiple layers and non-linearities (see Section  \ref{sec:extensions}), but leave this to future work. All training details can be found in the appendix (see Section \ref{appendix:experiment_details}).

In Figure \ref{fig:training_curve_2L} the plot on the right shows that the qualitative convergence properties of the approximate gradient methods are very similar to those of the exact gradient, eventually converging to nearly the same validation likelihood. However, as can be seen in the plot on the left, due to the reduced computational complexity, the approximate gradient method trains in less than half the time, and even more quickly to approximate convergence. This timing comparison is demonstrated more exactly in Table \ref{tab:timing_results_full} where the time per training batch and time per sample is computed for all models presented in this work. As can be seen, the self normalizing flow models are the fastest of all presented models, with the exception of the relative gradient method which appears to lag behind in likelihood performance. From this table we also see that the relative improvements to speed are directly related to the portions of the network which are replaced with self normalizing components. Since only the 1x1 convolution is replaced in the Glow framework, there is only a slight speed increase to be had. 

Finally, we quantitatively measure the quality of the gradient approximation by measuring the alignment of directions of the approximate gradient and the exact gradient in Figure \ref{fig:angle}. Specifically, for models trained in the self normalizing framework, we measure the `angular error' between the approximate gradient and the true gradient at each training iteration, for each layer, and plot the average angular error over all layers for each training epoch. The shaded area denotes the standard deviation in angles. We make two observations from this figure. First, both the 2-layer FC model and the 9-Layer CNN appear to have an initial slight divergence from the true gradient, but then quickly align to less than 1-degree of error, suggesting they are close approximations to the true gradient direction. Second, we observe that the average angular error of the approximate gradient is significantly larger for the convolutional model than it is for the fully connected model. When comparing this with the results in Table \ref{tab:nll_results} we see that this could be contributing to the slightly lower performance of the self normalizing model when compared with the exact gradient methods. We again  hypothesize this could be due to convolutional inverse constraint, making the inverse approximation more challenging. We leave further exploration of this topic to future work but believe that the performance could be ameliorated with improved inverse approximations, potentially achieved through more complex constrained optimization techniques.

\begin{table}
    \centering
    \caption{Bits per dimension for large scale Glow models on two larger scale natural image datasets. Mean $\pm$ std. over 3 runs. Self normalizing flows (SNF) achieve comparable performance to their exact counterparts demonstrating that this method scales to large models ($>100$ steps of flow). See Section \ref{appendix:experiment_details} for details.}
    \vspace{2mm}
    \begin{tabular}{l r r r}
    \toprule
        Model & CIFAR-10 & ImageNet32 \\ \midrule
        Glow & 3.36 $\pm$ 0.002 & 4.12 $\pm$ 0.002 \\
        SNF Glow & 3.37 $\pm$ 0.004 & 4.14 $\pm$ 0.007 \\
    \bottomrule
    \end{tabular}
    \label{tab:glow_results}
     \vspace{-3mm}
\end{table}

\begin{table*}[h]
    \centering
    \caption{Runtime comparison for the models presented in Tables \ref{tab:nll_results} and \ref{tab:glow_results}. 
    Hardware and implementation details are in Section \ref{appendix:experiment_details}} 
    \vspace{2mm}
    \label{tab:timing_results_full}
    \begin{tabular}{l l c c}
    \toprule
        Model & Dataset & Time / batch (ms) & Time / sample (ms) \\ \midrule
        Exact FC 2-Layer & MNIST & 44.9 $\pm$ 4.4 & 61.5 $\pm$ 5.8\\ 
        Relative Gradient FC 2-Layer & MNIST & 7.0 $\pm$ 0.4 & 69.2 $\pm$ 5.6 \\ 
        Self Normalizing FC  2-Layer & MNIST  & 18.7 $\pm$ 0.8 & 38.6 $\pm$ 3.1\\ \midrule
        Exact Conv. 9-Layer & MNIST  & 372.2 $\pm$ 24.5 & 241.6 $\pm$ 12.9 \\
        Emerging Conv. 9-Layer  & MNIST  & 305.0 $\pm$ 14.5 & 71.7 $\pm$ 8.8\\ 
        Conv. Exponential 9-Layer \ & MNIST & 304.4 $\pm$ 11.9 & 84.2 $\pm$ 9.2  \\ 
        Self Normalizing Conv. 9-Layer & MNIST & 212.5 $\pm$ 37.3 & 29.9 $\pm$ 6.3 \\
        \midrule
        Glow 2L-16K & MNIST & 583.4 $\pm$ 21.2 & 163.1 $\pm$ 21.8 \\
        Self Normalizing Glow 2L-16K & MNIST & 476.4 $\pm$ 16.7 & 30.6 $\pm$ 2.2 \\
        \midrule
        Glow 3L-32K & CIFAR-10 & 1841.3 $\pm$ 85.4 & 126.3 $\pm$ 13.9 \\
        Self Normalizing Glow 3L-32K & CIFAR-10 & 1761.2 $\pm$ 104.5 & 97.8 $\pm$ 12.9 \\
        \midrule
        Glow 3L-48K & ImageNet 32x32 & 2397.2 $\pm$ 204.0 & 174.8 $\pm$ 16.7 \\
        Self Normalizing Glow 3L-48K & ImageNet 32x32 & 2047.9 $\pm$ 152.8 & 150.7 $\pm$ 20.8 \\
\bottomrule
    \end{tabular}
\end{table*}

\subsection{CIFAR-10 and ImageNet 32x32}
For large scale experiments, we incorporate self normalizing flow layers into the Glow framework and train the models on CIFAR-10 and Imagenet 32x32. Specifically, we use the same model architectures as those proposed in \citet{kingma2018glow}, with some slight changes to the optimization parameters as detailed in Section \ref{appendix:experiment_details}. 

We observe that the self normalizing flow models are able to achieve competitive bits per dimension on both datasets (as seen in Table \ref{tab:glow_results}), while simultaneously training slightly faster than their exact gradient counterparts (as seen in Table \ref{tab:timing_results_full}). Importantly, we experimented with values of $\lambda$ in the set $\{1, 10, 100, 1000\}$, and chose $\lambda=1000$ for all CIFAR-10 and Imagenet models due to increased stability during training. We observed a slight reduction in final likelihood performance as a result of such a large reconstruction weight and believe this is a factor in the performance gap between the self normalizing and exact gradient models. We believe that with more tuning, or with a dynamically weighted constrained optimization method such as that presented in \citet{constrained_opt}, the self normalizing model is likely to match the exact gradient model even more closely. Preliminary experiments in this direction are shown in Section \ref{sec:extended_results}.

Although it is clear that the glow framework is not the optimal setting for the application of self normalizing layers, given the determinant calculation of the 1x1 convolution is relatively quick to evaluate already, we present this work as a proof of scalability of our framework to models with greater than 100 steps of flow (as in the ImageNet 32x32 case), and to larger scale images.

\section{Discussion}
\label{sec:discussion}
We see that the above framework yields an efficient update rule for flow-based models which appears to perform similarly to the exact gradient while taking significantly less time to train. 
However, in its current form, this approach is limited in a number of its own ways.

First, as described in Section \ref{sec:inference}, the evaluation of the exact log-likelihood is still expensive, requiring the computation of the exact log Jacobian determinant of a general function class. However, once a model is trained, the Jacobian determinants of the linear transformations only need to be computed once, and can then be re-used for all future likelihood evaluation, effectively amortizing the cost. 

Second, there are no known optimization guarantees for our proposed model. Therefore, the model could converge to a sub-optimal trade-off between the negative log-likelihood and the reconstruction error, or even diverge if the inverse approximation is very poor. In practice, we observe that reconstruction error stays very low for most models when initialized properly, and final likelihood values are only marginally impacted by the choice of $\lambda$. 
In future work, we intend to explore the possibility of augmented Lagrangian methods, the Modified Differential Method of Multipliers \cite{constrained_opt} and other constrained optimization techniques, which could provide better convergence guarantees.


We note one of the biggest constraints of the models presented here is that the inverse of the forward function may not always be given by a function of the same class, as for convolution. Although not evaluated in this work, we note that the general framework proposed here would allow for more complex asymmetric self normalizing flow components and we intend to evaluate these in future work.

Finally, as with all current exact likelihood flow methods, the dimensionality of the representation must stay consistent throughout the depth of the flow. Currently, this problem has been approached with flow components such as coupling layers and variational data augmentation, however these methods are either restrictive in their design or add significant noise. This is clearly one of the greatest architectural constraints for existing normalizing flows and remains so in this work.


\section{Conclusion}
In summary, we introduce Self Normalizing Flows, a new method to efficiently optimize normalizing flow layers. The method approximates the gradient of the log Jacobian determinant using learned inverses, allowing for the training of otherwise intractable normalizing flow architectures. We demonstrate that our method performs competitively with other models in the literature while simultaneously providing faster training and sampling.

\bibliography{main}
\bibliographystyle{icml2021}

\appendix
\pagebreak
\section{Appendix}
\subsection{Self Normalizing Fully Connected Gradients}
\label{appendix:fc_derivation}

In this section we provide an extended derivation of Equations $\ref{grad_w_approx}$ and $\ref{grad_r_approx}$ including derivation of the reconstruction gradient. For the gradient with respect to $\mW$, as given in Equation \ref{grad_w_approx}, we see that we only need to approximate the gradient of the Jacobian determinant term to achieve an efficient update, yielding:
\normalsize
\begin{align}
    \label{appendix_eqn:fc_grad_w}
    \frac{\partial}{\partial  \mW} \mathcal{L}(\vx) & \approx \frac{1}{2}\left(\bm\delta_{\vz}^{f} \vx^T + \mR^{T}\right) - \lambda \frac{\partial}{\partial \mW}  ||\mR \mW \vx - \vx||^2_2 \nonumber\\
    & = \frac{1}{2}\left(\bm\delta_{\vz}^{f} \vx^T + \mR^{T}\right) - 2 \lambda \mR^T(\mR\mW \vx - \vx) \vx^T
\end{align}
\normalsize
For the gradient with respect to $\mR$, we start with the exact gradient as given in Equation \ref{fc_grad_r}:
\begin{multline}
    \frac{\partial}{\partial  \mR} \mathcal{L}(\vx) = \frac{1}{2}  \left(-\mR^{-T} \bm\delta_{\vz}^{g} \vx^T \mR^{-T} - \mR^{-T}\right) \\ - \lambda \frac{\partial}{\partial \mR}  ||\mR \mW \vx - \vx||^2_2 
\label{appendix_eqn:fc_grad_r}
\end{multline}
\normalsize
To find an efficient approximation of this update without having to compute matrix inverses, we note that in addition to substituting $\mW$ in the place of $\mR^{-1}$, we also must approximate $\bm\delta_{\vz}^{g}$ in order to avoid having to compute $\log p_{\mZ}(\mR^{-1}\vx)$. We propose this can similarly be approximated as $\bm\delta_{\vz}^{g} \approx \bm\delta_{\vz}^{f} = \frac{\partial \log p_{\mZ}(\mW \vx)}{\partial \mW\vx}$ under the same constraint that $\mR^{-1} \approx \mW$. Together this yields:
\begin{multline}
    \frac{\partial}{\partial  \mR} \mathcal{L}(\vx)  \approx \frac{1}{2} \!\! \left(-\mW^{T} \bm\delta_{\vz}^{f} (\mW \vx)^T - \mW^{T}\right) \\ \!-\! 2 \lambda (\mR\mW \vx - \vx) (\mW\vx)^T
\label{appendix_eqn:fc_grad_r_cont}
\end{multline}
Conveniently, $\mW^T \delta^f_{\vz} = \frac{\partial \log p_{\mZ}(\mW \vx)}{\partial \vx}$ is then the delta from the output backpropagated to the input of the layer, and is computed by standard backpropagation. We call this term $\bm\delta^f_{\vx}$, giving a simplified gradient: 
\begin{equation}
    \label{appendix_eqn:fc_grad_r_final}
    \frac{\partial}{\partial  \mR} \mathcal{L}(\vx) \approx \frac{1}{2} \left(-\bm\delta^{f}_{\vx} \vz^T - \mW^{T}\right) - 2 \lambda (\mR\mW \vx - \vx)\vz^T 
\end{equation}
We see then that both Equations \ref{appendix_eqn:fc_grad_w} and \ref{appendix_eqn:fc_grad_r_final} require no matrix inverses, and furthermore require no additional terms beyond those computed by standard backpropagation. 

\subsection{Self Normalizing Convolution Gradients}
\label{appendix:conv_derivation}
In this section we provide a detailed derivation of the approximate gradient of the log Jacobian determinant term for convolutional layers. From Equations \ref{appendix:approx_conv_w} and \ref{appendix:exact_conv_r}, we see that after substitution of the approximate inverse we have:
\normalsize
\begin{align}
\label{appendix_eqn:approx_conv_w_cont}
    \frac{\partial}{\partial \vw} \log p^{f}_{\mX}(\vx) & \approx \bm\delta^f_{\vz} \star \vx + \frac{\partial (\mathrm{vec}\ \mathcal{T}(\vw))^T}{\partial \vw} \left(\mathrm{vec}\ \mathcal{T}(\vr)^{T}\right)
\end{align}
\begin{align}
\label{appendix_eqn:approx_conv_r_cont}
    \frac{\partial}{\partial \vr} \log p^{g}_{\mX}(\vx) & \approx - \bm\delta^f_{\vx} \star \vz - \frac{\partial (\mathrm{vec}\ \mathcal{T}(\vr))^T}{\partial \vr} \left(\mathrm{vec}\ \mathcal{T}(\vw)^{T}\right)
\end{align}
\normalsize
Focusing on the second term, we consider what $\frac{\partial (\mathrm{vec}\ \mathcal{T}(\vk))^T}{\partial \vk} \left(\mathrm{vec}\ \mathcal{T}(\vp)^{T}\right)$ is for arbitrary kernels $\vk, \vp \in \mathbb{R}^m$, and matrices $\mathcal{T}(\vk), \mathcal{T}(\vp) \in \mathbb{R}^{D \times D}$. First, we see $\frac{\partial (\mathrm{vec}\ \mathcal{T}(\vk))^T}{\partial \vk} \in \mathbb{R}^{m \times D^2}$, meaning that it is a rectangular matrix where each row corresponds to one dimension of the kernel, and each column corresponds to one element of the matrix $\mathcal{T}(\vk)$. Writing out the partial derivative element-wise, we get:
\begin{align}
    \left[\frac{\partial (\mathrm{vec}\ \mathcal{T}(\vk))^T}{\partial \vk}\right]_{i,j}  = \frac{\partial \left[\mathrm{vec}\ \mathcal{T}(\vk)\right]_j}{\partial k_i}
\end{align}
From this it is clear that the $i^{th}$ row and $j^{th}$ column will be $1$ if $\left[\mathrm{vec}\ \mathcal{T}(\vk)\right]_j = k_i$ and otherwise $0$. Intuitively, given the definiton of $\mathcal{T}(\vk)$, this corresponds to an indicator matrix, indicating whether kernel element $i$ is shared at location $j$ of the vectorized convolution matrix $\mathcal{T}(\vk)$. 

Given this intuition, we can see that the inner product of the $i^{th}$ row of this matrix and the same vectorized convolution matrix $\mathrm{vec}\ \mathcal{T}(\vk)$ will yield a sum over elements of $\mathcal{T}(\vk)$ where $\mathcal{T}(\vk) = k_i$. Formally:
\begin{align}
    \frac{\partial (\mathrm{vec}\ \mathcal{T}(\vk))^T}{\partial k_i} \mathrm{vec}\ \mathcal{T}(\vk) 
    & = \sum_{j}^{D \times D} k_i * \mathbbm{1}\left[\left[\mathrm{vec}\ \mathcal{T}(\vk)\right]_j = k_i\right] \nonumber \\
    & = m_i * k_i
\end{align}
where $m_i$ is the number of times the kernel element $k_i$ is shared across the matrix $\mathcal{T}(\vk)$.

Similarly, for the inner product between a row of this indicator matrix and a vectorized convolution matrix formed from a different kernel (i.e. $\mathcal{T}(\vp)$), we see that the result is again given by a multiple times the new kernel: 
\begin{align}
\label{appendix_eqn:mp}
    \frac{\partial (\mathrm{vec}\ \mathcal{T}(\vk))^T}{\partial k_i} \mathrm{vec}\ \mathcal{T}(\vp) 
    & = \sum_{j}^{D \times D} p_i * \mathbbm{1}\left[\left[\mathrm{vec}\ \mathcal{T}(\vk)\right]_j = k_i\right] \nonumber \\
    & = m_i * p_i
\end{align}
This can be understood to be due to the fact that the mapping $\mathcal{T}(\cdot)$ which generates the weight sharing structure is the same for both the partial derivative matrix and the vectorized convolution matrix. 

\paragraph{Computing the Transposed Convolution Kernel $\mathrm{flip}(\cdot)$\newline} 
In Equations \ref{appendix_eqn:approx_conv_w_cont} and \ref{appendix_eqn:approx_conv_r_cont} we see the partial derivative matrix is multiplied with a transposed convolution matrix. For the convolution operations proposed here, the transpose convolution matrix can also be written as a standard convolution matrix with a transformed kernel. We denote this transformation as $\mathrm{flip(\cdot)}$ such that $\mathcal{T}(\mathrm{flip}(\vk)) = \mathcal{T}(\vk)^T$, or equivalently $\mathrm{flip}(\vk) = \mathcal{T}^{-1}(\mathcal{T}(\vk)^T)$. This transformation can easily be implemented in deep learning frameworks through index manipulation of the kernel. Informally, this operation is achieved by swapping the input and output axes, and mirroring the spatial dimensions. Explicitly, given a four dimensional kernel $\vk \in \mathbb{R}^{O \times I \times H \times W}$ where $O, I, H, W$ are the number of output channels, number of input channels, kernel height, and kernel width respectively, the flip operation can be defined as:
\begin{align}
    \mathrm{flip}(\vk)_{o,i,h,w} = k_{i,o,H-h, W-w}
\end{align}

\paragraph{Computing the Multiple $\vm$ \newline}
The constant $\vm$ which is the same shape as the kernel, and is element-wise multiplied, is given by the number of times each element $k_i$ of the kernel $\vk$ is present in the matrix $\mathcal{T}(\vk)$. This can be easily computed as a convolution of two images filled entirely of $1's$, the first with the shape of the outputs, and the second with the shape of the inputs. Using syntax from the PyTorch framework, we can write this as:
\begin{equation}
    \vm = \mathrm{ones\_like}(\vz) \star \mathrm{ones\_like}(\vx)
\end{equation}
Note this convolution must be performed with the same parameters as the main convolution (e.g. padding, stride, grouping, dilation, etc.).

Combining Equation \ref{appendix_eqn:mp} with the $\mathrm{flip}$ operation, for all kernel elements $i$, we see that we arrive at Equations \ref{appendix:approx_conv_w_complete} and \ref{appendix:exact_conv_r}.

\subsection{Experiment Details}
All code for this paper can be found at the following repository: \url{https://github.com/AKAndykeller/SelfNormalizingFlows}
\label{appendix:experiment_details}

\paragraph{Training \& Evaluation Details\\}
In Tables \ref{tab:nll_results} and \ref{tab:glow_results}, all log-likelihood (in nats) and bits per dimension values are computed using the exact log Jacobian determinant of the full transformation. They are reported on a held-out test set, using the saved model parameters from the best epoch as determined by performance on a separate validation set. Each value is given as a mean $\pm$ standard deviation as computed over 3 runs with different random initializations.

For MNIST, the first 50,000 training images were used for training, and the last 10,000 were used for validation. The plots in figure \ref{fig:training_curve_2L}, show the negative log-likelihood on the MNIST validation set for the 2-layer fully connected (FC) models. 

For CIFAR-10, the first 40,000 training images were used for training and the last 10,000 were used as a held out validation set. Data augmentation including random horizontal flipping with probability 0.5 and random jitter by 1 pixel was performed to prevent overfitting.

For ImageNet 32x32, a random subset of 20,000 images were used for validation, and the remaining 1,261,149 images were used for training. The dataset was constructed using the same methodology as \citealt{kingma2018glow}, and can be downloaded from \url{http://image-net.org/small/download.php}. The  values reported in Table \ref{tab:glow_results} were computed on the provided 50,000 image test set. No data augmentation was performed. 

\paragraph{Optimization Details\\}
All fully connected (FC) models were trained for 6000 epochs using Adam optimizer \cite{adam} with a batch size of 100, a learning rate of $1\times10^{-4}$, $\beta_1=0.9, \beta_2=0.999$, and reconstruction weight $\lambda=1$. These parameters were chosen to match the training methodology of \cite{gresele2020relative}. The loss was computed as the average over the batch. All convolutional (Conv.) models were trained for 1000 epochs with the same optimization parameters, but with a learning rate of $1\times10^{-3}$ due to observed faster convergence. 

The MNIST Glow models were trained for 250 epochs using the same optimizer settings as the Conv. models. We experimented with values of $\lambda$ in the set $\{1, 10, 100, 1000\}$, and chose the $\lambda=100$ based on the highest validation accuracy. 

The CIFAR-10 Glow models were trained for 1000 epochs with the same optimizer settings, but with $\lambda=1000$. Additionally, the norm of the gradients was clipped at $10,000$ for the self normalizing CIFAR-10 models for improved stability during training. 

The ImageNet 32x32 Glow models were trained for 15 epochs, with the same optimization parameters and again with  $\lambda=1000$. The norm of the gradients was clipped at $10,000$ for all ImageNet models.

All models except for the ImageNet models were trained using a learning-rate warm-up schedule where the learning rate is linearly increased from 0 to its full value over the course of the first 10 epochs.

\paragraph{Discrepancies with Published Results\\}
We note that our value for the Relative Gradient model \cite{gresele2020relative} differs from the published result of $-1375.2 \pm 1.4$. We found experimentally that when using the same settings as published in \cite{gresele2020relative}, our re-implementation achieved approximately $-1102$. We found  the discrepancy to be due almost exactly to the log Jacobian determinant of the data-preprocessing steps (such as dequantization, normalization, and the logit transform), which we measure to sum to $272.7 \pm 0.3$. We discussed with the authors of  \cite{gresele2020relative} and concluded they likely did not include the log Jacobian determinant of these steps in their reported values. We further note the numbers in Table $\ref{tab:nll_results}$ are using slightly different parameters than in \cite{gresele2020relative}, such as $\alpha=0.3$ in the activation function, a batch size of 100, and a significantly longer training duration.

We additionally see that our values in Table \ref{tab:glow_results} are slightly worse than those reported in \cite{kingma2018glow} (i.e. 3.36 vs. 3.35 on CIFAR-10, and 4.12 vs 4.09 on ImageNet 32x32). We hypothesize that our slightly worse performance is likely due to our use of an explicit validation set, reducing the effective size of the training set. To the best of our knowledge, this pre-processing step was not performed in \cite{kingma2018glow}. Additional factors could be a shorter training duration, no learning rate warmup (for ImageNet), or the imposed gradient clipping.

\paragraph{Timing details\\}
\label{appendix:timing_details}
\looseness=-1
Figure \ref{fig:timing} was created by running a single fully connected layer (with no activation function) on a machine with an NVIDIA GeForce 1080Ti GPU and Intel Xeon E5-2630 v3 CPU. Each datapoint was computed by taking the mean and standard deviation of the time required per batch over 4,000 batches, with a batch size of 128, on synthetically generated random data at integer multiples of 96 dimensions starting at 32. 

The values reported for MNIST in Table \ref{tab:timing_results_full} were computed on the same machine with an NVIDIA GeForce 1080Ti GPU and Intel Xeon E5-2630 v3 CPU, with a batch size of 100. The values for CIFAR-10 and ImageNet were computed on a machine with an NVIDIA Titan X GPU and Intel Xeon E5-2640 v4 CPU, with batch sizes of 100 and 64 respectively. The discrepancy between training and sampling time for the FC models is due to the iterative optimization required to invert the Smooth Leaky ReLU activation. Figure \ref{fig:training_curve_2L} was created using the time results from Table \ref{tab:timing_results_full}. For all times reported, the times of the first and last 100 batches per epoch were ignored to reduce variance. 

\paragraph{Architectures\\}
\label{appendix:architectures}
All models were trained using pre-processed data in the same as manner as \cite{gresele2020relative, maf, real_nvp}. This includes uniform dequantization, normalization, and logit-transformation. We additionally use a standard Gaussian as our base distribution $p_{\mZ}$ for all models. 

\looseness=-1
All 2-layer fully connected (FC) models use the Smooth Leaky ReLU (with $\alpha=0.3$) activation (as in \cite{gresele2020relative}). Weights of the forward model ($\mW$'s) are initialized to identity plus noise drawn from a Xavier Normal \cite{xavier} with gain $0.01$. Weights of the inverse model ($\mR$'s) are initialized to the transpose of the forward weights.

All 9-layer convolutional models are trained with spline \cite{splines} activations with individual parameters per pixel and 5-knots, kernels of size $(3\times3)$, and zero-padding of 1 on all sides. The convolutional models are additionally divided into three blocks, each of 3 layers, with 2 `squeeze' layers in-between the blocks. The squeeze layers move feature map activations from the spatial dimensions into the channel dimension, reducing spatial dimensions by a half and increasing the number of channels by 4 (as in \cite{hoogeboom2019emerging}). Weights of the forward model ($\vw$'s) are initialized with the dirac delta function (preserving the identity of the inputs) plus noise drawn from a Xavier Normal \cite{xavier} with gain $0.01$. Weights of the inverse model ($\vr$'s) are initialized to $\mathrm{flip}(\vw)$.

The Glow models for MNIST were constructed of $L=2$ blocks of $K=16$ steps each (as specified in \cite{kingma2018glow}), where each block is composed of a squeeze layer and K-steps of flow. A split layer is placed between the two blocks. Each step of flow is composed of an act-norm layer, a $(1 \times 1)$ convolution, and an affine coupling layer. The coupling layers are constructed as in \cite{kingma2018glow}. All convolutional weights were initialized to random orthogonal matrices. For CIFAR-10 and ImageNet 32x32, the Glow models were composed of $L=3, K=32$ and $L=3, K=48$ respectively, matching those in \cite{kingma2018glow}.

\subsection{Proposed Model Extensions}
\label{sec:extensions}

\paragraph{Asymmetric Convolutions} 
As mentioned in Section \ref{sec:discussion}, the inverse of the forward function may not always be given by a function of the same class (e.g. for convolutional layers). To partially alleviate this constraint, we propose that the forward and inverse functions may be asymmetric, and derive a simple case of this below for convolutions with different kernel sizes (but identical output sizes). Initial experiments with such an asymmetric model (i.e. 3$\times$3 conv. $f$ with 7$\times$7 conv. $g$) have shown promising results – improving over models with 3$\times$3 convolutions in both directions. 

For a function $f$ with a (column vector) kernel $\vw \in \mathbb{R}^{m}$ , and an inverse $g$ with a larger kernel $\vr \in \mathbb{R}^n$ ($n > m$), we see that the approximate gradient with respect to $\vw$ can be obtained by taking the internal central dimensions of $\vr$, and similarly the gradient for $\vr$ is given by taking a zero-padded version of $\vw$. Formally we can write the central-indexing and padding operations as multiplication by the rectangular matrices $\mP_{\vr\rightarrow\vw}$ and $\mP_{\vw\rightarrow\vr}$ respectively. 
\begin{equation}
\label{appendix:approx_conv_w_fc_g}
\begin{split}
    \frac{\partial}{\partial \vw} \log p^{f}_{\mX}(\vx)  \approx \bm\delta^f_{\vz} \star \vx + \mathrm{flip}(\mP_{\vr\rightarrow\vw}\vr) \odot \vm\\
\end{split}
\end{equation}
\begin{equation}
\label{appendix:approx_conv_w_fc_g_Rgrad}
\begin{split}
    \frac{\partial}{\partial \vr} \log p^{g}_{\mX}(\vx) & \approx - \bm\delta^f_{\vx} \star \vz - \mathrm{flip}(\mP_{\vw\rightarrow\vr}\vw) \odot \vm 
\end{split}
\end{equation}

Where $\mP_{\vr\rightarrow\vw}$ and $\mP_{\vw\rightarrow\vr}$ are defined as:
\begin{equation}
\mP_{\vr\rightarrow\vw} = \left[\begin{matrix}
    \mathbf{0} & \mathbf{I}_{m} & \mathbf{0}
\end{matrix}\right] \ \ \ \ \ \& \ \ \ \ \ \  \mP_{\vw\rightarrow\vr} = \left[\begin{matrix}
    \mathbf{0} \\ \mathbf{I}_{n} \\ \mathbf{0}
\end{matrix}\right]
\end{equation}
Where $\mathbf{I}_D$ refers to a $D \times D$ identity matrix, and $\mathbf{0}$ is a matrix of zeros such that the dimensions of $\mP_{\vr\rightarrow\vw}$ are $m \times n$ and those of $\mP_{\vw\rightarrow\vr}$ are $n \times m$.

\paragraph{Jacobian Vector Product Inverse Constraint}
As noted in Section \ref{sec:general_framework}, the approximations made assume that the Jacobians of the functions $f$ and $g$ are approximately inverses in addition to the functions themselves being approximate inverses. As stated, for the models presented in this work, this property is obtained for free since the Jacobian of a linear mapping is the matrix representation of the map itself. However, since this property may not hold in general, we propose the following additional constraint could be added to the objective:
\begin{equation}
    \mathcal{E}_{JVP}(f, g) = - \mathbb{E}_{\mathbf{\nu} \sim \mathcal{U}(0,1)} \left[ ||\mJ_{g} \mJ_{f} \mathbf{\nu} - \mathbf{\nu}||^2_2 \right]
\end{equation}
where $\mJ_g, \mJ_f$ are the Jacobians of $g$ and $f$ respectively, evaluated at $\vz$ and $\vx$ respectively. The expectation can additionally be approximated by Monte Carlo methods through a finite number of samples.

We see that such a loss would reach a minimum when the Jacobians are exact inverses. However, it remains unclear from which distribution the points $\mathbf{\nu}$ should be sampled to achieve the best approximate inverse. Since Jacobian vector products are efficiently implemented in most deep learning frameworks, (and are thus much faster than naive matrix-matrix multiplication) this loss could be added to the overall objective while still avoiding $\mathcal{O}(D^3)$ computational complexity.

\paragraph{Variational Interpretation}
One limitation of the self normalizing flow framework becomes apparent mainly in the presence of data-dependant transformations, as would be contained in the general framework outlined in Section \ref{sec:general_framework}. Specifically, in this setting, the invertibility of the transformation is more difficult to guarantee since the value of the Jacobian determinant is then a function of the input, and therefore ensuring invertibility is not as simple as computing non-zero determinants on the training data.

To alleviate this difficulty, we believe a variational interpretation of the self normalizing framework as afforded by \citet{flowvae} would relax the global invertibility constraint to a local invertibility constraint. To achieve this, the encoder and decoder become stochastic with a very small noise level $\sigma^2$, and in the limit of $\sigma^2 \rightarrow 0$, the variational lower bound becomes equal to change of variables formula. In such a model, the inverse approximation error would contribute to the decoder likelihood directly, and the `self normalizing' inverse approximation would become useful in computing the gradient of the entropy of the encoder. As the authors note, such an interpretation allows for non-invertible encoders, and we thus believe it is a fruitful direction for future research.

\subsection{Extended Results}
\label{sec:extended_results}
\paragraph{Choice of $\lambda$}
As noted in the discussion, we observe that final likelihood values are only marginally impacted by the choice of $\lambda$. To quantitatively demonstrate this, we present the performance for different reconstruction weights $\lambda$ in Table \ref{tab:different_lambdas} below. As can be seen, for values of $\lambda$ greater than a minimum threshold, the likelihood performance decreases slightly as $\lambda$ increases. For values of $\lambda$ which are too small, training is initially stable but eventually becomes unstable and diverges, resulting in numerical instability (denoted  `-').
\begin{table}[h!]
    \centering
    \vspace{-4mm}
    \caption{NLL in nats on MNIST for SNF models with different values of $\lambda$. `-' implies the model training diverged.}
    \label{tab:different_lambdas}
    \vspace{2mm}
    \begin{tabular}{l r r r r r r r}
    \toprule
    \tikz[diag text/.style={inner sep=0pt, font=\footnotesize},
      shorten/.style={shorten <=#1,shorten >=#1}]{%
        \node[below left=2pt and 28pt, diag text] (def) {\hspace{-4pt}Model};
        \node[above right=-2pt and -7pt, diag text] (abc) {$\hspace{-20pt}\lambda$};
        \draw[shorten=4pt, very thin] (def.north west|-abc.north west) -- (def.south east-|abc.south east);} 
         & \hspace{-6mm}$\leq$ 0.1 & 1.0 & 10.0 & 100.0 & 1000.0 \\ \midrule
         2L FC & \hspace{-6mm}- & 946.3 & 954.6 & 1007.4 & 1039.8 \\
         9L Conv. & \hspace{-6mm}- & 639.6 & 642.9 & 646.3 & 660.6 \\
         2L-16K Glow &\hspace{-6mm} - & - & - & 574.1 & 574.8 \\
    \bottomrule
    \end{tabular}
     \vspace{-4mm}
\end{table}

\paragraph{Improved Constrained Optimization}
In this work, to enforce the approximate-inverse constraint $f^{-1} \approx g$, we make use of a penalty method with parameter $\lambda$ (the `reconstruction weight'). The downsides of this method are that it requires manual tuning of $\lambda$ which can lead to sub-optimal local minima (as seen above). A powerful alternative to the basic penalty method is the method of Lagrange multipliers, whereby $\lambda$ is simultaneously optimized with the model parameters by a min-max optimization scheme. One related implementation of such a method is given by the GECO algorithm from \citet{rezende2018taming}. Simply, the update equation of the algorithm is given by  $\lambda^{t} \leftarrow \lambda^{t-1} \mathrm{exp}(\propto C^t)$ for each iteration $t$, where $C^t$ is derived from the exponential moving average of the constraint (i.e. reconstruction loss). In initial experiments, we have shown such a method works  well when combined with the self normalizing framework, reducing the need for tuning, and enabling the stable training of more complex functions (see Table \ref{table:geco} below). The implementation of this method is additionally provided in the code repository.

\begin{table}[h!]
\label{table:geco}
    \centering
    \vspace{-4mm}
    \caption{NLL in nats on MNIST comparing the impact of larger kernel sizes incorporated into the Glow framework (2L-4K width=16). SNF models are trained with the method of Lagrange multipliers. Mean $\pm$ std  over 3 random initalizations.}
    \vspace{2mm}
    \begin{tabular}{l r r r r}
    \toprule
        Model & Glow 1x1 & SNF 1x1 & SNF 5x5\\ \midrule
         \hspace{-4pt}$-\log p_{\mathbf{X}}(\mathbf{x})$ & 678.3 $\pm$ 2.0 & 678.3 $\pm$ 9.7 & 669.9 $\pm$ 2.8 \\
    \bottomrule
    \end{tabular}
     \vspace{-2mm}
\end{table}

\subsection{Novelty Comparison}
We provide Table \ref{tab:invertibleComparison} (adopted from \cite{behrmann2019iresnet}), to facilitate comparison of the self normalizing framework with existing normalizing flow methods.  

\begin{table*}[h!]
\begin{center}
\caption{High level comparison of Self Normalizing Flows (SNF) with existing normalizing flow frameworks.\vspace{5mm}}

\begin{tabular}{c|ccccc|c}
      \toprule
      {Method} & {NICE/ i-RevNet} & {Real-NVP} & {Glow} & {FFJORD} & {i-ResNet} & SNF \\ \midrule
      {Free-form}  & {\xmark} & {\xmark} & {\xmark} & {\cmark} & {\cmark} & {\cmark} \\
      {Analytic Forward}  & {\cmark} & {\cmark} & {\cmark} & {\xmark} & {\cmark} & {\cmark} \\ 
      {Analytic Inverse} & {\cmark}& {\cmark} & {\xmark} & {\xmark} & {\xmark}  & {\xmark} \\
      {Non-volume Preserving} & {\xmark} & {\cmark} & {\cmark} & {\cmark} & {\cmark} & {\cmark} \\
      {Exact Likelihood}  & {\cmark} & {\cmark} & {\cmark} & {\xmark} & {\xmark}  & {\xmark} \\
      {Unbiased Stochastic Log-Det Estimator} & {N/A} & {N/A} & {N/A} & {\cmark} & {\xmark}  & {\xmark} \\
      {Unconstrained Lipschitz Constant} & {\cmark} & {\cmark} & {\cmark} & {\cmark} & {\xmark}  & {\cmark} \\
      \bottomrule
\end{tabular}
\label{tab:invertibleComparison}
\end{center}
\end{table*}

\subsection{Acknowledgements}
We would like to thank the creators of Weight \& Biases \cite{wandb} and PyTorch \cite{pytorch}. Without these tools our work would not have been possible. We thank the Bosch Center for Artificial Intelligence for funding, and Anna Khoreva and Karen Ullrich for guidance. Finally, we thank the reviewers for their proposed extensions and constructive comments.

\end{document}